\documentclass[10pt,twocolumn,letterpaper]{article}

\usepackage[applications]{wacv}      %

\usepackage{graphicx}
\usepackage{amsmath}
\usepackage{booktabs}
\usepackage{multirow}
\usepackage{algpseudocode}
\usepackage{algorithm}
\usepackage{amssymb}

\usepackage[pagebackref,breaklinks,colorlinks]{hyperref}

\usepackage[capitalize]{cleveref}
\crefname{section}{Sec.}{Secs.}
\Crefname{section}{Section}{Sections}
\Crefname{table}{Table}{Tables}
\crefname{table}{Tab.}{Tabs.}

\def\wacvPaperID{1961} %
\def\confName{WACV}
\def\confYear{2024}

\begin{document}

\title{Towards Diverse and Consistent Typography Generation}

\author{Wataru Shimoda$^{1}$ \qquad Daichi Haraguchi$^{2}$ \qquad Seiichi Uchida$^{2}$ \qquad Kota Yamaguchi$^{1}$ \\
$^{1}$CyberAgent, Japan \qquad $^{2}$Kyushu University, Japan\\
}

\maketitle

\begin{figure*}
  \centering
  \includegraphics[width=1.\textwidth]{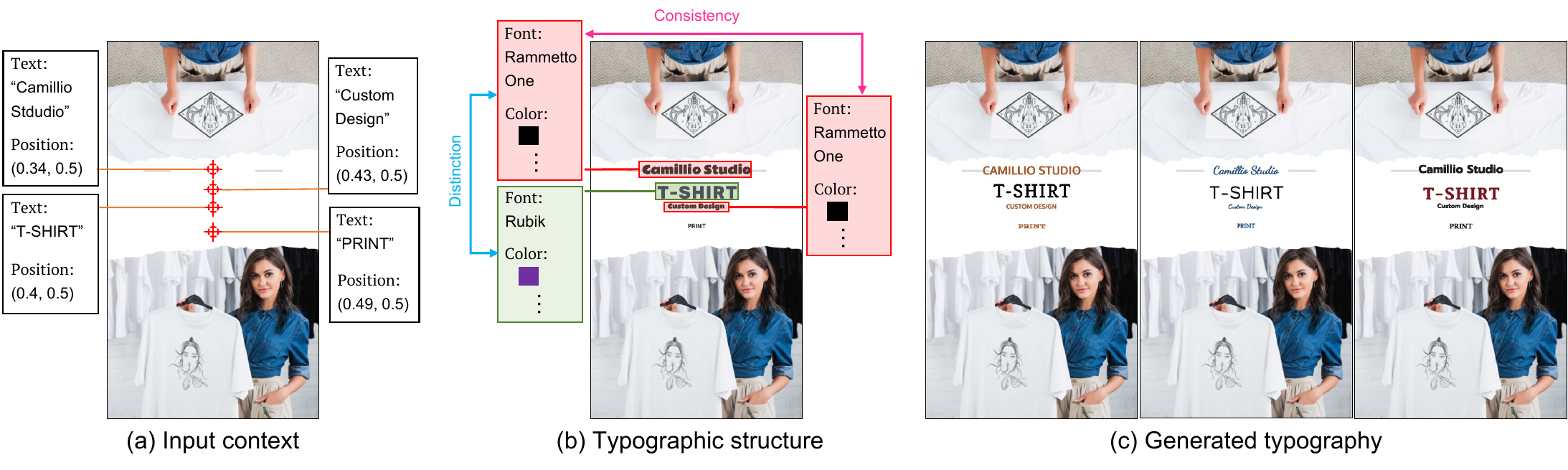}
  \caption{We formulate the fine-grained typography generation task considering the structure of multiple texts.
  a)~An example of an input context: background image, texts, and their corresponding center positions.
  b)~Typographic structure predicted by our model via top-1 sampling.
  c)~Generated typography by our structure-preserved sampling.
  }
  \label{fig:teaser}
\end{figure*}

\begin{abstract}
In this work, we consider the typography generation task that aims at producing diverse typographic styling for the given graphic document.
We formulate typography generation as a fine-grained attribute generation for multiple text elements and build an autoregressive model to generate diverse typography that matches the input design context.
We further propose a simple yet effective sampling approach that respects the consistency and distinction principle of typography so that generated examples share consistent typographic styling across text elements.
Our empirical study shows that our model successfully generates diverse typographic designs while preserving a consistent typographic structure.
\end{abstract}

\section{Introduction} \label{sec:intro}
In textual communication, typographers carefully express their intent in their typographic work, such as product packages, posters, banner ads, book covers, signboards, and presentation slides.
Appropriately designed typography affects how people perceive the impression, legibility, and importance of the text content, yet choosing appropriate typography is surprisingly challenging~\cite{Bringhurst1992}.
Typographic design involves a complex interplay between the message content, background visuals, layout arrangement, and styling consistency across text elements.

In building a practical automatic typography system, we have to take into account the following requirements.

\noindent\emph{Context awareness}: A system should reflect the context of the creative work; e.g., styling should emphasize the word ``Sale'' for a sale event poster or use serif-style fonts with careful letter spacing for luxury brands to express their authority.
Also, typography should match the background visuals; e.g., a bright font color for a dark background.

\noindent\emph{Fine-grained representation}: A system can handle fine-grained typographic attributes beyond font family and color, such as horizontal text alignment, line spacing, letter spacing, or angle, that are important to convey a delicate nuance within the graphic design.

\noindent\emph{Consistency and distinction}: A system should apply consistent style across multiple texts that share the same semantics~\cite{robin2014non-designer}; e.g., menu items should have uniform styling.
On the other hand, typography should have distinct styling to emphasize the content semantics; e.g., a title should be highlighted by a different font family and size.

\noindent\emph{Diversity}: A system should be able to suggest diverse design candidates to the users because there is usually no single optimal typographic design in a real-world creative workflow.

In this paper, we formulate the typography generation task as fine-grained typographic attribute generation and build an autoregressive model that can generate diverse yet consistent typographic styling for the given graphic document.
Given a canvas, texts, and their rough positions (Fig.~\ref{fig:teaser}a), our model generates fine-grained attributes such as font, color, or letter spacing for each text element.

Our model relies on the attention mechanism of the Transformer architecture to capture the consistency-relationship among texts as well as the relationship between texts and the input context.
For generating diverse typography, we propose a simple yet effective sampling approach to enforce consistent styling among text elements, which we refer to as \emph{structure-preserved sampling}.
Our sampling approach predicts which text elements share the uniform styling in the first step (Fig.~\ref{fig:teaser}b) and samples diverse attributes constrained by the predicted relationships in the second step (Fig.~\ref{fig:teaser}c).
We also propose metrics to evaluate the quality of typography generation, where we define the typography structure in the form of pairwise consistency relationships among text elements.

We show in experiments that our autoregressive models outperform baseline approaches and successfully generate diverse typography that respects context and consistency.
Our user study also confirms that our approach is qualitatively preferred over the baseline.
Our attribute-based formulation is readily applicable in a real-world creative workflow, as designers usually work on graphic documents with vector-graphic authoring tools like Adobe Illustrator.

We summarize our main contributions in the following.
\begin{itemize}%
\item We formulate the typography generation task that aims at jointly generating diverse fine-grained typographic attributes.
\item We present an autoregressive approach to generate typographic attributes, where we develop the structure-preserved sampling to generate diverse yet consistent typographic designs.
\item We propose metrics to evaluate the quality of typography generation that is aware of the consistency among text elements.
\item We empirically show the effectiveness of our approach both quantitatively and qualitatively.
\end{itemize}
\section{Related work} \label{sec:related_work}

\subsection{Attribute-based typography generation}
While attribute-based representation is commonly observed in commercial design authoring tools, we do not find much literature on attribute-based typography generation.
MFC~\cite{fontsincontext} is a notable exception that predicts the font, color, and font size of a single text box from the global image, local image, and auxiliary tag information.
AutoPoster~\cite{autoposter} recently proposes a poster generation approach that also considers font, color, and font size within the model.
While the previous work considers typographic attributes, we consider far more fine-grained attributes including text angle, alignment, letter spacing, and line spacing, and explicitly consider consistency relationships among multiple text elements.
Other notable works include the study of Jiang \etal~\cite{fontpair} on combinatorial preference in font selection for subjects and subtitles in PDF data and Shimoda \etal~\cite{Shimoda_2021_ICCV} proposing a de-rendering approach to parse rendering parameters from texts in raster images.

\subsection{Raster typography generation}
Raster typography generators directly render stylized texts in pixels.
There are two types of formulations: text style transfer and conditional stylized text generator.
Text style transfer aims at generating stylized text images for the specified styles.
Awesome Typography is a style transfer method by a patch matching algorithm~\cite{awesometypography}.
Recent literature reports several GAN-based models~\cite{mcgan, tetgan, uceffect,iccvtextst,typographydecor,dyntypo, glyph_style_fewshot}.
Wang \etal propose a layout-specified text style transfer method~\cite{textlogosynthesis}.
Raster text editing is another branch of the text style transfer task, where the goal is to apply a reference style to the manually edited image~\cite{srnet,stefann,swaptext}.

There are several neural network-based glyph renderers without reference images.
We refer to these approaches as conditional stylized text generators.
Miyazono \etal and Gao \etal~\cite{textpainter} propose a generative model that directly produces stylized texts in the raster format from background images, layouts, and text contents~\cite{miyazono}.
Recent text-to-image model~\cite{dalle, stable_diffusion} can draw stylized texts via prompts, but these models tend to corrupt glyphs in the raster format~\cite{sd_text}.
Some recent works propose fine-tuned text-to-image models~\cite{glyphdraw, diffste, glyphcontrol} that address glyph corruption.

While there are quite a few works on raster generation, attribute-based generation has a clear practical advantage in that the generation result is 1) free from raster artifacts and 2) easily applicable in real-world authoring tools.

\subsection{Graphic design generation}
Our typography generation task can be regarded as one sub-topic within the broader study of attribute-based graphic design or layout generation.
Early work on layout generation utilizes templates~\cite{template_layout0, template_layout1} or heuristic rules~\cite{heuristic_layout0}.
Recent literature relies on neural networks for generation.
LayoutVAE~\cite{LayoutVAE} generates scene layouts from label sets using autoregressive VAE.
LayoutGAN~\cite{LayoutGAN} adopts a GAN-based layout generator via a differentiable wire-frame model.
VTN~\cite{Arroyo_2021_CVPR}, LayoutTransformer~\cite{gupta2021layouttransformer}, and CanvasVAE~\cite{canvasvae} report Transformer-based VAE for graphic designs.
LayoutDM~\cite{inoue2023layoutdm} adopts a discrete diffusion model to layout generation.
Towards finer control on the generation quality, several literature~\cite{Content-aware-GModel, Neural-Design-Network, Kikuchi2021, Patil_2020_CVPR_Workshops, kong2022blt, layout_imgcondition, layout_imgcondition2, layout_system, inoue2023layoutdm, jiang2022layoutformer, blt} tackles to generate layouts with constraining and conditional information.
While most recent attempts seem to be interested in the layout-level generation, our focus is the unique and explicit modeling of text styling in the typographic design.

\begin{table}[tb]
    \centering
    \footnotesize
    \caption{Context and typographic attributes. Context attributes consist of canvas input and element input.}
    \vspace{-3mm}
    \label{tab:context}
    \begin{tabular}{cccc}
      \hline
      Type & Name & Modality & Size\\
      \hline
      \multirow{3}{*}{\begin{tabular}{c}Canvas\\input \\$\mathbf{x}_\mathrm{canvas}$\end{tabular}}
      & Background & Image & \small $256 \times 256 \times 3$\\
      & Aspect ratio & Categorical & 40 \\
      & Number of text & Categorical & 50\\
      \hline
      \multirow{6}{*}{\begin{tabular}{c}Element\\input\\$\mathbf{x}_\mathrm{t}$\end{tabular}}
      & Text & Text & variable\\
      & Left & Categorical & 64\\
      & Top & Categorical & 64\\
      & Line count & Categorical & 50\\
      & Char count & Categorical & 50\\
      & Background & Image & \small $256 \times 256 \times 3$\\
      \hline
      \multirow{8}{*}{\begin{tabular}{c}Typographic\\attributes\\(output) \\$\mathbf{y}_\mathrm{t}$\end{tabular}}
      & Font & Categorical & 261\\
      & Color & Categorical & 64\\
      & Alignment & Categorical & 3\\
      & Capitalization & Categorical & 2\\
      & Font size & Categorical & 16\\
      & Angle & Categorical & 16\\
      & Letter spacing & Categorical & 16\\
      & Line spacing & Categorical & 16\\
      \hline
    \end{tabular}
\end{table}
\section{Approach}  \label{sec:problem}
Our goal is to generate typography with consistency and diversity from context attributes such as \emph{background image}, \emph{texts}, and their corresponding \emph{center positions}.
To this end, our model first predicts typographic structure (Fig.~\ref{fig:teaser}b) and then generates typography through a structure-preserved sampling of typographic attributes such as \emph{font} and \emph{color} (Fig.~\ref{fig:teaser}c).

\subsection{Problem formulation}

We define the context attributes by $X \equiv (\mathbf{x}_\mathrm{canvas}, \mathbf{x}_1, \dots, \mathbf{x}_T)$, where $\mathbf{x}_\mathrm{canvas} \equiv (x_\mathrm{background}, x_\mathrm{aspect}, \dots)$ denotes a tuple of canvas input and $\mathbf{x}_t \equiv (x_\mathrm{text}^t, x_\mathrm{top}^t, x_\mathrm{left}^t, \dots)$ denotes the $t$-th element input.
We assume there are $T$ text elements in the document.
We consider target typographic attributes $Y \equiv (\mathbf{y}_1, \dots, \mathbf{y}_T)$, where $\mathbf{y}_t \equiv (y_\mathrm{font}^t, y_\mathrm{color}^t, \dots)$ is typographic attributes of the $t$-th text element. 
Our goal is to generate typographic attributes $Y$ by a conditional generation model $p_\theta$ parametrized by $\theta$:
\begin{align}
    \hat{Y} & \sim p_\theta(Y | X). \label{eq:form}
\end{align}

\subsection{Typographic attributes}  \label{sec:typography}
Our context and typographic attributes contain multiple modalities, which we preprocess into feature representation beforehand.
We summarize the feature representation of all attributes in Table~\ref{tab:context}.
Our context attributes consist of the canvas input and the element input.
We extract a background image for both the global canvas and the region of each text element, resize the image to the fixed resolution with the RGB format, and finally apply an ImageNet pre-trained ResNet50~\cite{resnet} to extract features.
We preprocess text content using a pre-trained CLIP encoder~\cite{clip}.
We discretize continuous attributes, such as an aspect ratio or a position, based on the k-means clustering, where we empirically set the appropriate number of clusters.

\begin{figure}[t]
\centering
\includegraphics[width=0.67\columnwidth]{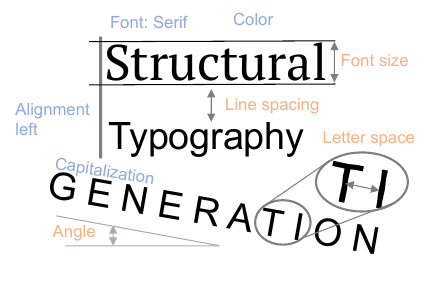}\\[-4mm]
\caption{An illustration of typographic attributes.
We handle semantic quantities including \emph{font}, \emph{color}, \emph{alignment}, and \emph{capitalization} and geometric quantities including \emph{font size}, \emph{angle}, \emph{letter spacing}, and \emph{line spacing}.
}
\label{fig:attribute}
\end{figure}

In this work, we consider the following typographic attributes as outputs: \emph{font}, \emph{color}, \emph{font size}, \emph{alignment}, \emph{caplitalization}, \emph{angle}, \emph{letter spacing}, and \emph{line spacing} for each text element.
Our typographic attributes have semantic and geometric quantities.
We show the illustration of the typographic attributes in Fig.~\ref{fig:attribute}.
We also discretize typographic attributes based on the k-means clustering.

\begin{figure}[t]
\centering
\includegraphics[width=0.9\columnwidth]{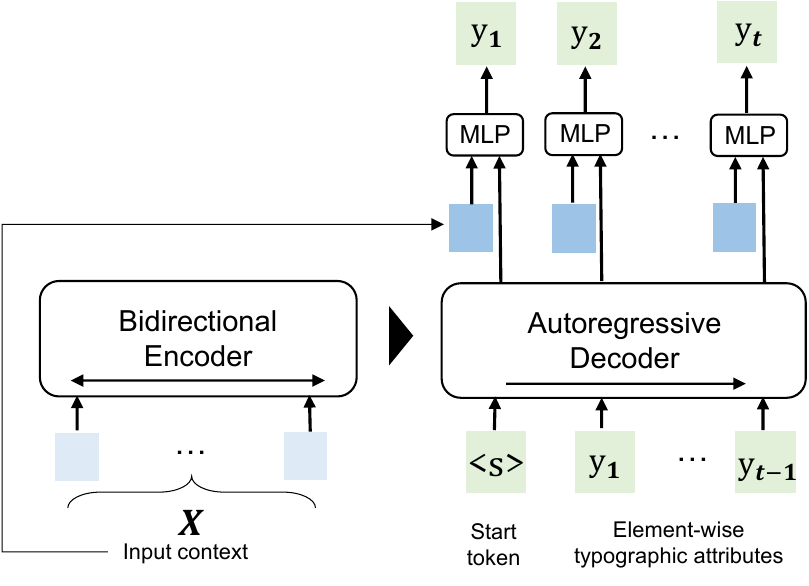}\\[-2mm]
\caption{Model architecture.}
\label{fig:model}
\end{figure}

\begin{figure}[t]
\centering
\includegraphics[width=\columnwidth]{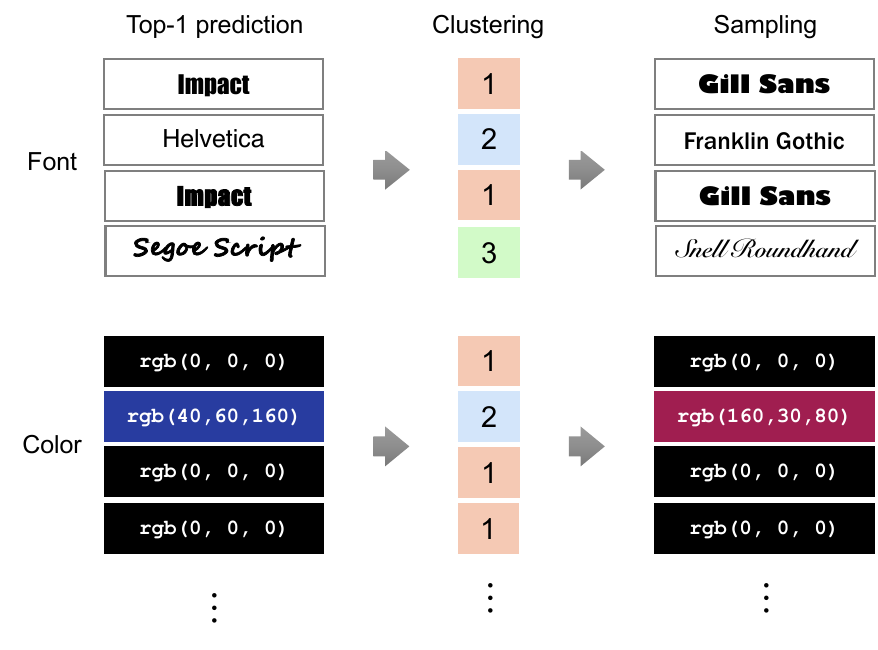}\\[-3mm]
\caption{Our structure-preserved sampling first clusters elements by top-1 prediction, then draws samples per cluster so that the result maintains the most likely typographic structure while being capable of generating various design.}
\label{fig:structured_sampling}
\end{figure}
\subsection{Typography generation}  \label{sec:arch}
We build an encoder-decoder architecture based on Transformer~\cite{transformer} to effectively capture the interaction among the inputs and the target attributes within the attention mechanism.
Fig.~\ref{fig:model} illustrates the overall architecture.
Our architecture combines BART-style Transformer blocks~\cite{bart} with skip connections between the input and output of each element.
We project input features into fixed-size embeddings and feed into the Transformer encoder blocks.

We adopt an autoregressive decoder to model the joint distribution of typographic attributes:
\begin{align}
    p_\theta(Y | X) = \prod_t^T p_\theta(\mathbf{y}_t | \mathbf{y}_{t-1}, \dots, \mathbf{y}_1 , X),
\end{align}
and we apply element-wise autoregressive sampling to generate attribute $k$ at the $t$-th element:
\begin{align}
    \hat{y}_{k}^t \sim p_\theta(y_k^t | \mathbf{y}_{t-1}, \dots, \mathbf{y}_1, X). \label{eq:sampling}
\end{align}
Here, we apply top-$p$ sampling~\cite{top-p} to draw attributes.
Top-$p$ sampling has a hyper-parameter $p_k \in [0, 1]$ that controls the diversity for each attribute $k$.
In our experiments, we fix $p_k=0.1$ for geometric attributes (font size, angle, letter spacing, and line spacing) to avoid visually disturbing generation, and vary $p_k$ for other attributes depending on the experimental setup.

To train the model, we minimize the following objective:
\begin{align}
    \sum_t \sum_k \mathcal{L}_\mathrm{entropy}^k(y_k^t, \tilde{y}_k^t) + \lambda_\mathrm{reg} |\theta|^2,
\end{align}
where $\mathcal{L}_\mathrm{entropy}^k$ is the standard cross entropy for the attribute $k$, $\tilde{y}_k^t$ is the ground truth, and $\lambda_\mathrm{reg}$ is the L2 weight decay.

\subsection{Structure-preserved sampling} \label{sec:sampling}

While autoregressive sampling can adjust sampling hyper-parameter for each attribute, we find the plain autoregressive approach sometimes corrupts the consistency and distinction among element styling (Sec.~\ref{sec:intro}), especially when we increase the parameter $p_k$ of top-$p$ sampling.
Here, we propose the \emph{structure-preserved sampling}, which is a simple two-step inference approach that effectively controls the diversity while preserving the typography structure.
The general steps are the following.
\begin{enumerate}
    \item Infer initial prediction $\check{Y}$ via top-1 sampling:
\begin{align}
    \check{y}_k^{t} = \mathrm{argmax} ~ p_\theta (y_k^t | \mathbf{y}_{t-1}, \dots, \mathbf{y}_1, X).  \label{eq:prediction}
\end{align}
    \item For each attribute $k$, cluster text elements $\mathcal{T} \equiv \{1, \dots, T\}$ by label linkage $\check{y}_k^{t} = \check{y}_k^{t'}$ for any pair $t \neq t'$.
    \item Autoregressively sample $\hat{y}_k^t$ again but assign the same label if any element in the same cluster is already assigned a label.
\end{enumerate}
In both inference steps, we keep the same raster scan order of elements (left-to-right, top-to-bottom).
Basically, we autoregressively sample over clusters instead of all the elements in the second sampling step.
Fig.~\ref{fig:structured_sampling} illustrates the above steps.
The intuition is that top-1 sampling gives the best typographic structure, and the second sampling generates diverse examples while forcing the consistent structure from the initial inference.
Our approach is heuristic but generates visually plausible typography without significant overhead.

It is possible to replace the initial top-1 sampling with other sampling approaches if we need to generate a typographic design with a different structure.
In this work, we assume a typical typographic design does not require a diverse structure in the application scenario; e.g., design suggestion in an authoring tool.

We split the clustering step for each attribute, but it is also possible to consider joint clustering across attributes.
The challenge here is that a different attribute has a different perception in the final visualization.
It is not straightforward to define a unified cluster affinity across typographic attributes; e.g., humans perceive the difference in a font more than the different alignments.
In our dataset, we often observe texts that share the same font but with different sizes.
We leave the optimal design of typographic clusters for our future work.

\section{Evaluation Metrics} \label{sec:evaluation_metrics}
There is no standardized evaluation metric for typography generation.
We adopt several metrics to evaluate typography generation performance.

\subsection{Attribute metrics}
In our setting, we handle several typographic attributes, but the format of each attribute is not the same.
Here, we introduce evaluation metrics for measuring the fidelity of attribute prediction.

\noindent\textbf{Accuracy:}
We evaluate categorical attributes (\emph{font}, \emph{align}, \emph{capitalization}) by the standard accuracy metric between the prediction and the ground truth.

\noindent\textbf{Mean absolute error:}
We evaluate the geometric attributes by the absolute difference in their respective unit.
We measure \emph{font size} in points, \emph{angle} in degree, \emph{letter spacing} in points, and \emph{line spacing} in a relative scale centered at 1.0.

\noindent\textbf{Color difference:}
We employ CIEDE2000 color difference~\cite{ciede2000} to measure the similarity between colors, which is known to well reflect the human perception of color difference.

\subsection{Structure score}

The structure score examines whether the use of the same attribute pairs matches the ground truth.
That is, if a pair of texts share the same attribute, we assign \texttt{1}, and if the pair differs, we assign \texttt{0}, then measure the accuracy between the prediction and the truth.
Formally, for attribute $k$, we consider the set of binary labels over any pair of text elements:
\begin{align}
    S_k(Y) &\equiv \{ \delta(y_k^i, y_k^j) | i \in \mathcal{T}, j \in \mathcal{T}, i \neq j \},
\end{align}
where $\delta(y_k^i, y_k^j)$ is an indicator function that measures the condition $y_k^i = y_k^j$.
The structure score is the accuracy of prediction $S_k(\hat{Y})$ against the ground truth $S_k(Y)$ for each document.

\subsection{Diversity score}

We evaluate how diverse the generated typography attributes are.
Assuming we generate $N$ samples, we count the average number of unique labels over elements in the generated samples:
\begin{align}
    \frac{1}{T} \sum_{t}^T \frac{N_{\mathrm{uniq}, k}^t}{N},
\end{align}
where $N_{\mathrm{uniq}, k}^t$ is the unique count of attribute $k$ at the $t$-th element.

\section{Experiments} \label{sec:experiments}
We evaluate typography generation performance as well as top-1 prediction performance for fair comparison.

\subsection{Dataset} \label{sec:datasets}
We evaluate the generation task using the Crello dataset~\cite{canvasvae}, which includes various design templates in vector format.
Since the original dataset does not contain all of the necessary typographic information for visualization, we collect additional resources like \texttt{ttf} files.
We parsed and compiled the typographic details of each template, and finally obtained 23,475 templates that contain text elements in the vector format.
We split the Crello dataset to {\it train}:{\it test}:{\it val} with an 8:1:1 ratio, (i.e., 18,780, 2,347, 2,347).

\subsection{Implementation details} \label{sec:details}
We set the dimension of feature embeddings to 256.
We set the feed-forward dimension to 512 and the number of the head to 8 in the Transformer blocks.
We stack 8 Transformer blocks in our model.
We use AdamW~\cite{adamw} optimizer with a 0.0002 learning rate and 30 epochs to train our model.

\begin{table*}[t]
  \centering
\caption{Attribute metrics. Acc: accuracy, MAE: mean absolute error, and Diff: CIEDE2000 color difference.} \label{tab:element_fidelity}
\vspace{-3mm}
  \begin{tabular}{lcccccccc}
    \hline
    \multirow{2}{*}{Method}
    & Font 
    & Color 
    & Align 
    & Capitalize
    & Size 
    & Angle 
    & Letter space
    & Line height
    \\
    & \small{Acc (\%) $\uparrow$}
    & \small{Diff (-) $\downarrow$}
    & \small{Acc (\%) $\uparrow$}
    & \small{Acc (\%) $\uparrow$}
    & \small{MAE (pt) $\downarrow$}
    & \small{MAE (°) $\downarrow$}
    & \small{MAE (pt) $\downarrow$}
    & \small{MAE (-) $\downarrow$}
    \\
    \hline
 Mode & 16.6 & 53.2 & 91.9 & 54.1 & 45.1 & 0.30 & 2.31 & 0.102 \\
 MFC & 
 10.4 \footnotesize{±6.87} & 54.9\footnotesize{±5.06}  & - & - & 28.0\footnotesize{±8.82} & - & - & - \\
 CanvasVAE* & 
 27.7\footnotesize{±9.56} & 53.3\footnotesize{±1.41} & 92.3\footnotesize{±0.43} & 66.0\footnotesize{±9.07} & 32.5\footnotesize{±3.97} & 0.30\footnotesize{±0.01} & 2.23\footnotesize{±0.09} & 0.095\footnotesize{±0.006} \\
    Ours & 
    40.9\footnotesize{±0.76} & 53.7\footnotesize{±1.96} & 93.8\footnotesize{±0.74} & 75.3\footnotesize{±0.67} & 20.9\footnotesize{±0.66} & 0.26\footnotesize{±0.02} & 2.16\footnotesize{±0.16} & 0.065\footnotesize{±0.003} \\
    \hline
  \end{tabular}
\end{table*}

\begin{table*}[t]
  \centering
\caption{Structure scores (\%).} \label{tab:structure_fidelity}
\vspace{-3mm}
  \begin{tabular}{lcccccccc}
    \hline
    Method
    & Font 
    & Color
    & Align
    & Capitalize
    & Font size 
    & Angle 
    & Letter space
    & Line height
    \\
    \hline
 Mode & 61.9 & 58.0 & 66.8 & 85.7 & 22.4 & 83.5 & 57.4 & 56.6
 \\
 MFC & 59.8\footnotesize{±4.22} & 58.9\footnotesize{±3.40} & - & - & 63.4\footnotesize{±6.19} & - & - & - \\
 CanvasVAE* & 
 62.0\footnotesize{±0.56} & 59.6\footnotesize{±1.84} & 66.5\footnotesize{±0.59} & 85.7\footnotesize{±0.22} & 43.7\footnotesize{±18.40}& 83.9\footnotesize{±0.42} & 58.7\footnotesize{±1.48} & 60.1\footnotesize{±3.80} \\
    Ours & 
    68.6\footnotesize{±0.44} & 66.9\footnotesize{±0.65} & 68.1\footnotesize{±0.58} & 86.3\footnotesize{±0.55} & 71.3\footnotesize{±0.55}& 86.0\footnotesize{±0.37} & 63.8\footnotesize{±0.77} & 78.9\footnotesize{±1.06} \\
    \hline
  \end{tabular}
\end{table*}

\begin{figure*}[t]
\centering
\includegraphics[width=1.\textwidth]{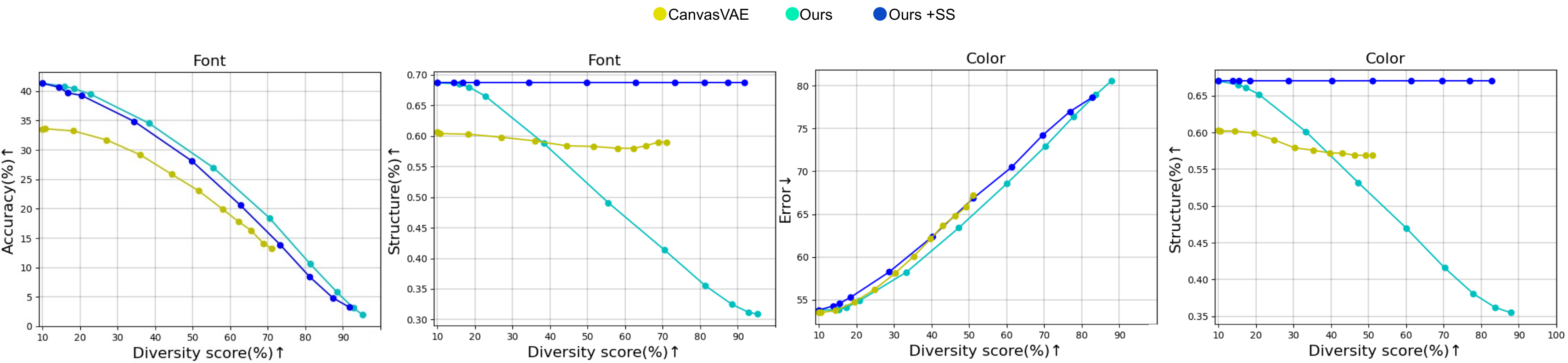}\\[-2mm]
\caption{Generation performance in terms of attribute metrics vs. diversity score for font and color attributes.
Our models outperform the CanvasVAE baseline by a large margin.
Our structure-preserved sampling further keeps the constant structure score regardless of the sampling parameter $p_k$.}
\label{fig:plots}
\end{figure*}
\begin{figure*}[t]
\centering
  \includegraphics[width=\textwidth]{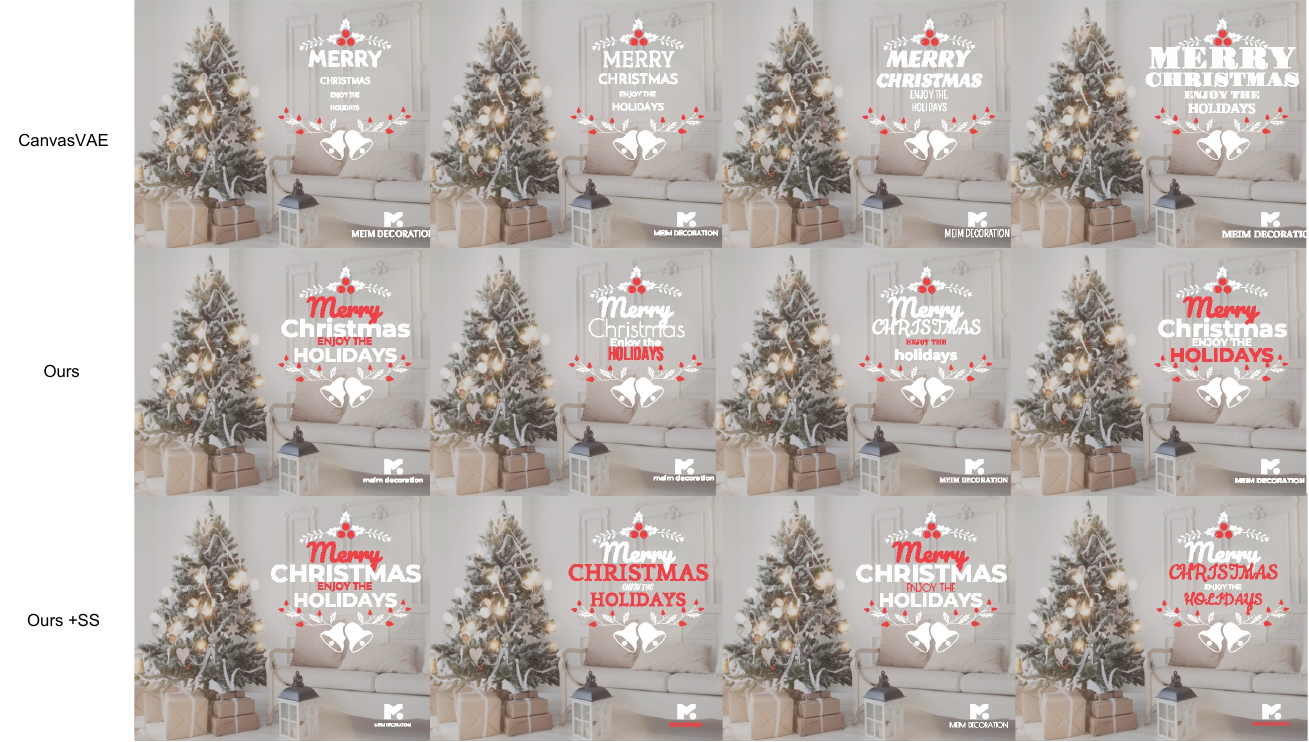}\\[-2mm]
\caption{Qualitative comparison of typography generation.
Our models generate sufficiently diverse typography with appropriate color tones to the background.
With the structure-preserved sampling, our model further enforces consistent styling like fonts to multiple texts (Ours+SS).}
\label{fig:gencmp}
\end{figure*}

\begin{figure}[t]
\centering
  \includegraphics[width=\columnwidth]{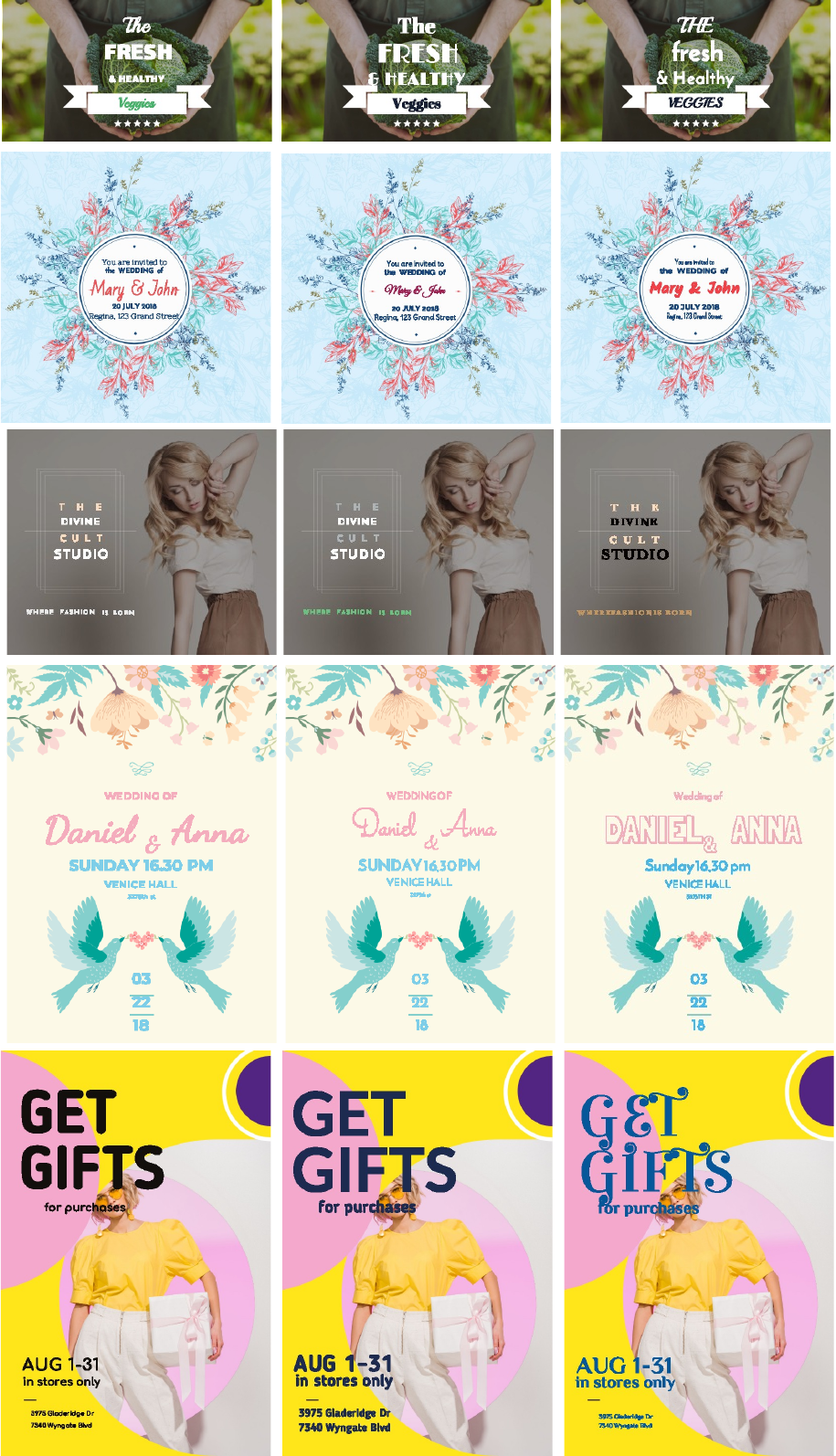}
\caption{Diverse generation examples. Each row shows three generated examples for the same input.}
\label{fig:genex}
\end{figure}

\subsection{Prediction evaluation}  \label{sec:baseline_comparison}
Here, we evaluate the performance of the top-1 prediction for a fair comparison with the previous work.
We compare the following baselines.

\noindent\textbf{Mode} always predicts the most frequent category, which shows the bias of each attribute in the dataset.

\noindent\textbf{MFC}~\cite{fontsincontext} is a fill-in-the-single-blank model tailored for typography.
This model predicts three attributes: \emph{font}, \emph{font size}, and \emph{color}.
MFC learns to predict embedding for font representation by minimizing L2 loss and adversarial loss, a scalar value for font size by minimizing the L1 loss, and a discretized token for color.
The embedding for font representation is obtained by a simple autoencoder.
Since this model cannot produce multiple outputs, we repeatedly apply the model to generate multiple outputs in an autoregressive manner.
We do not consider external contexts (HTML tags and design tags) used in \cite{fontsincontext} since the Crello dataset does not contain such resources. 

\noindent\textbf{CanvasVAE*}~\cite{canvasvae} is a Transformer-based variational autoencoder model for structured elements, including layout and canvas information~\cite{canvasvae}.
Since CanvasVAE is an unconditional model, we adapt CanvasVAE to accept input contexts and predict typographic attributes.
For the prediction task, we fix the bottleneck latent of the VAE to the mean vector.

\noindent\textbf{Ours} is the initial autoregressive prediction of our model (Sec~\ref{sec:problem}).

Table~\ref{tab:element_fidelity} and Table~\ref{tab:structure_fidelity} summarize the quantitative prediction performance.
Our model achieves the best scores in all structure scores, though not always the best in attribute metrics.
Interestingly, while our model shows moderate improvement over baselines in attribute metrics like \emph{font size}, our model shows significant improvement in terms of the structure score.
We observe that our model outperforms MFC even if MFC designs a dedicated loss for each attribute.
Our model also outperforms CanvasVAE, perhaps because CanvasVAE has a limited model capacity due to the global latent that is regularized to follow the normal distribution.
In distinction, our autoregressive models have sufficient capacity to model rich conditions across attributes and elements.

\subsection{Generation evaluation} \label{sec:generation_eval}
We generate 10 samples for each test input for evaluation.
We compare the following baselines.

\noindent\textbf{CanvasVAE*} is the same model we evaluate in Sec~\ref{sec:baseline_comparison}.
We control the generation diversity by scaling the coefficient of standard deviation in the latent space.

\noindent\textbf{Ours} is our model with a plain top-$p$ sampling and without our structure-preserved sampling.
We control the generation diversity by the hyper-parameter $p_k \in [0, 1]$ of top-p sampling except for geometric attributes.

\noindent\textbf{Ours+SS} applies the structure-preserved sampling to the above model.

\paragraph{Quantitative results} \label{sec:quantitative}

Fig.~\ref{fig:plots} plots the attribute metrics and the structure score of font and color as we increase the diversity hyper-parameter.
We observe that our models show a good quality-diversity trade-off compared to CanvasVAE.
While a plain top-p approach clearly degrades the structure score as we increase the diversity, our structure-preserved sampling keeps the constant score in the highly diverse regime.
Note that our structure-preserved sampling can slightly drop the attribute metrics compared to the plain autoregressive sampling due to the cases when the initial structure prediction fails.

\paragraph{Qualitative results} \label{sec:qualitative}
Fig.~\ref{fig:gencmp} shows qualitative results.
We set the diversity hyper-parameter of CanvasVAE to $std=100$, Ours to $p=0.9999$, and Ours+SS to $p_k=0.99999$, which yields similar diversity scores.
CanvasVAE tends to ignore the input context.
We suspect CanvasVAE suffers from learning a good single latent space for a complex task like typography generation.
Besides, CanvasVAE cannot independently control the diversity of different attributes, which causes poor overall appearance.
Our models generate sufficiently diverse typography for individual attributes in each element, and with the structure-preserved sampling, the results hold consistent styling across elements.
We show more generation examples by our model in Fig.~\ref{fig:genex}.
The first row, the second row, and the third row show examples that have only a few elements but have sufficient contrast.
The fourth and fifth rows show that our model consistently generates diverse yet plausible typography even when a document has many text elements.

\begin{figure}[t]
\centering
  \includegraphics[width=\columnwidth]{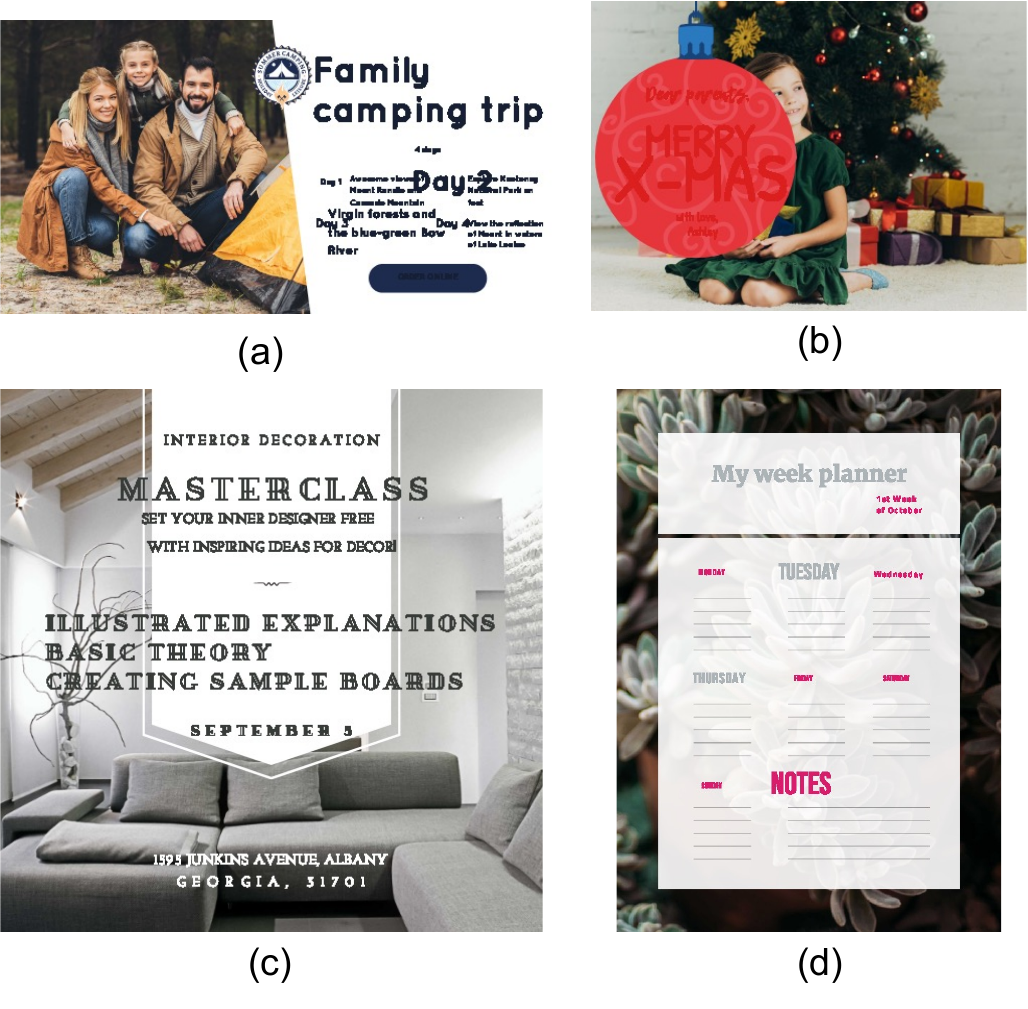}\\[-3mm]
\caption{Failure cases.}
\label{fig:failurecases}
\end{figure}

\paragraph{Limitation}
We show some failure cases of our approach in Fig.~\ref{fig:failurecases}.
Our model does not explicitly handle the appearance of typography and sometimes generates unintentional spatial overlaps between texts (Fig.~\ref{fig:failurecases}a), colors that are difficult to see
(Fig.~\ref{fig:failurecases}b), and overflow of a text element due to the unawareness of the final text width (Fig.~\ref{fig:failurecases}c).
Further, if our model fails to capture plausible structure, generated results corrupt (Fig.~\ref{fig:failurecases}d).

\subsection{User study}\label{sec:userstudy}
To verify that our evaluation metrics accurately reflect human perception, we conducted pilot user studies.
We asked ten participants to choose which generated design groups they preferred in a pairwise comparison between the two methods.
We compared the generation quality of our model with the CanvasVAE and our model without the structure-preserved sampling.
Each user study comprises 100 questions, resulting in 1000 responses in total.

As the diversity hyper-parameter affects generation quality, we choose the hyper-parameter of each approach to be comparable.
Specifically, we set the diversity hyper-parameter to have the diversity score within 49.8-51.5\% for font and 33.3-35.2\% for color in the CanvasVAE comparison,
and the diversity score within 70.4-73.3\% for the font and 60.0-61.3\% for the color in the plain sampling baseline.
We pick the diversity scores from Fig.~\ref{fig:plots}.

\begin{figure}[t]
\centering
\includegraphics[width=0.9\columnwidth]{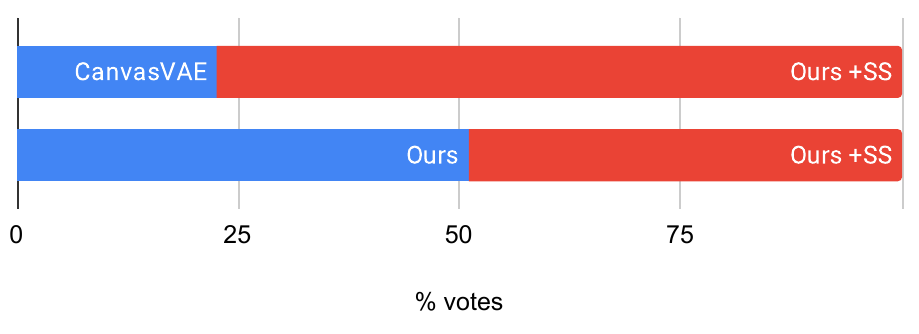}\\[-3mm]
\caption{User preference in pairwise comparison.}
\label{fig:userstudy}
\end{figure}

Fig.~\ref{fig:userstudy} summarizes the user preference.
We confirm that participants clearly prefer our model compared to CanvasVAE.
The results support the hypothesis that our quantitative results indeed reflect human perception.
On the other hand, our structure-preserved sampling does not make a difference in user preference.
While unexpected, we suspect that our sampling hyper-parameter was too diverse to give appropriate colors to texts and that made the pairwise comparison difficult for users.
In the future, we wish to continue on studying how to suggest the most comfortable designs.

\section{Conclusion}
In this paper, we formulate the task of typography generation where we have to generate diverse yet compelling typography given the input contexts.
We build a fine-grained typographic attribute generation model and propose a sampling technique to generate diverse typography with consistency and distinction among texts.
The empirical study confirms our approach successfully generates diverse yet consistent typography and outperforms the baselines.

There are remaining research questions we wish to explore.
We hope to analyze the relationship between attributes to human perception, as we identify that the fidelity of colors to the given background somehow dominates the first impression of the design.
We also hope to study to what degree of diversity users prefer in the generated results, for building a practical typography generation system.

{\small
\bibliographystyle{ieee_fullname}
\bibliography{main}
}

\cleardoublepage
\begin{center}
\textbf{\LARGE Supplemental Materials}
\end{center}
\section{Details of typographic attributes}
\noindent\emph{Font}:
We categorize fonts based on font family name.
We do not model the similarity of fonts in this paper.

\noindent\emph{Color}:
The color attribute is the color of the texts for filling.

\noindent\emph{Alignment}:
If texts have line breaks, the alignment attribute aligns texts in left, right, or center. 

\noindent\emph{Capitalization}:
The capitalization attribute is a binary option whether to capitalize texts or not.

\noindent\emph{Font size}:
The font size attribute is an input parameter of font, and it controls the size of texts.

\noindent\emph{Angle}:
The angle of texts for rotation.

\noindent\emph{Letter spacing}:
The letter spacing attribute represents the distance of letters in texts.

\noindent\emph{Line spacing}:
The line spacing attribute is a scale of interval in lines.

\section{Dataset statistics}
\begin{figure*}[t]
  \centering
  \includegraphics[width=1.\textwidth]{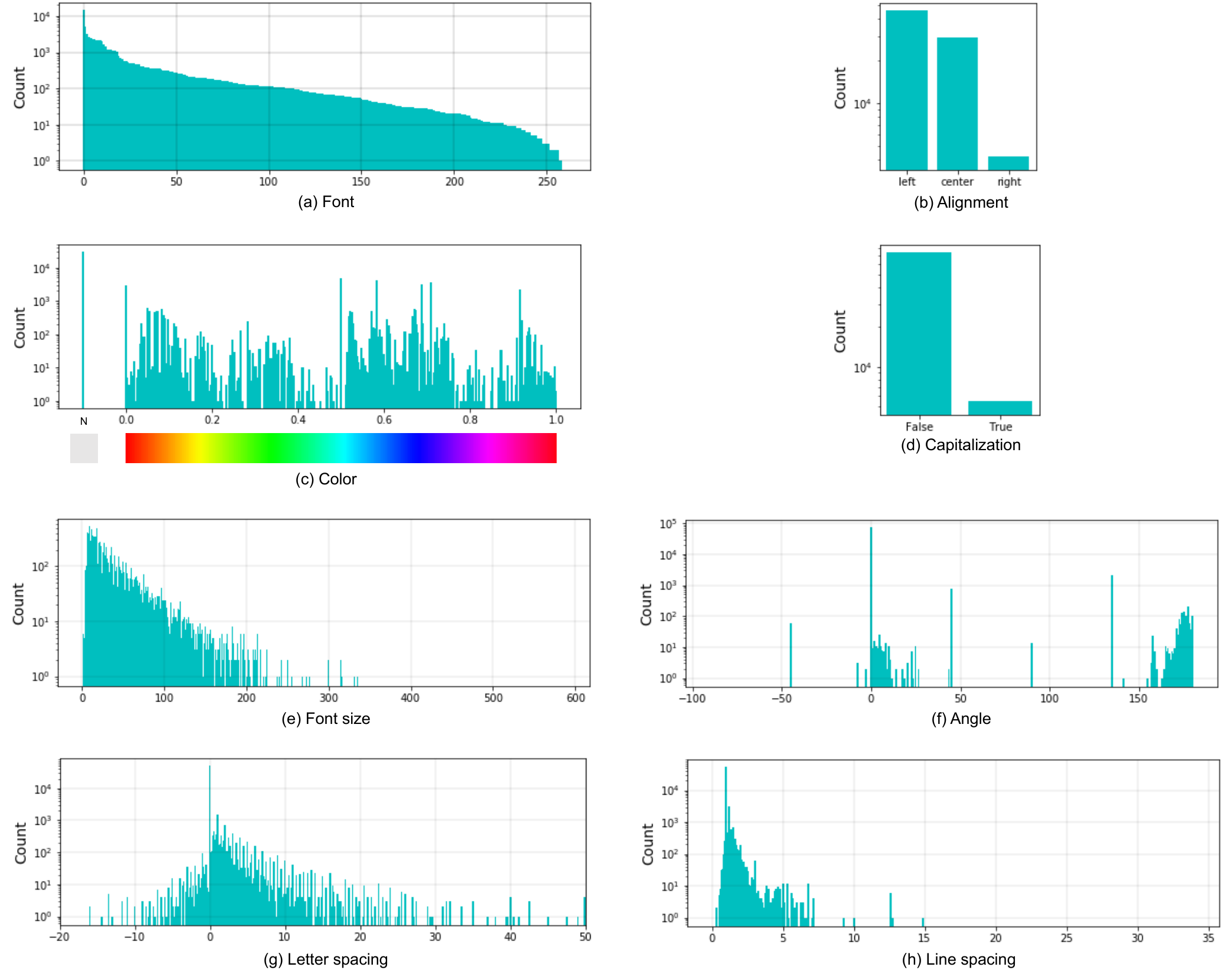}
\caption{The distributions of typographic attributes in the Crello dataset. The y-axis of plots is the logarithmic scale. N in the color distribution represents the neutral color. The (a) to (d) are semantic attributes, and (e) to (h) are geometric attributes.}
  \label{fig:dist}
\end{figure*}

We show statistics of typographic attributes of the Crello dataset~\cite{canvasvae}.
Fig.~\ref{fig:dist} shows the distribution.
We observe strong biases in typographic attributes that designers prefer to use.

\begin{figure*}[t]
  \centering
  \includegraphics[width=1.0\textwidth]{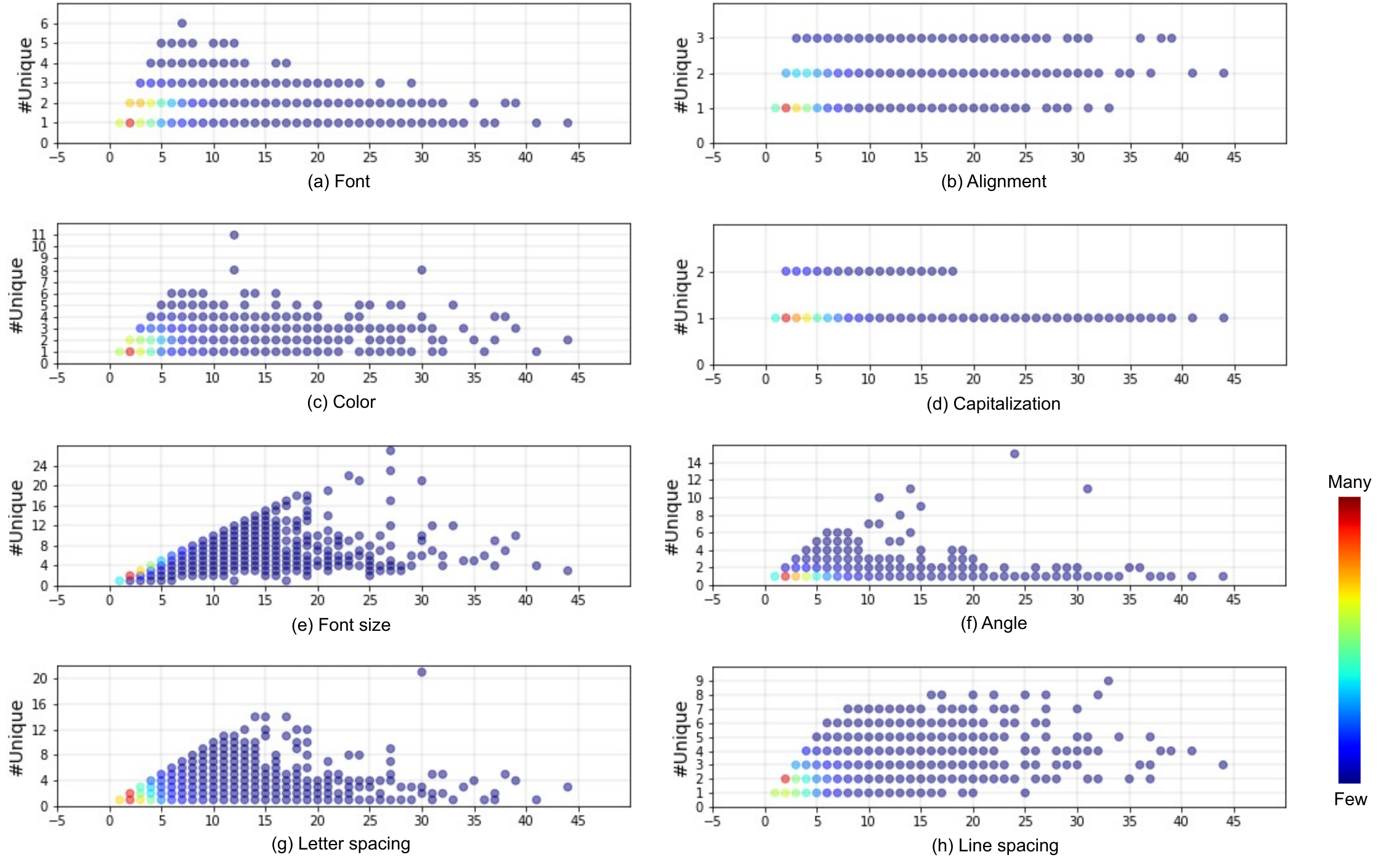}
  \caption{The use of typographic attributes by the number of text elements. 
  The color represents the count of unique labels.}
  \label{fig:unique}
\end{figure*}

Fig.~\ref{fig:unique} shows the number of unique labels in typographic attributes per design in the Crello dataset.
Even if there are many text elements, there are only a few attributes in use, and we rarely observe more than three different fonts in a single design document.
Geometric attributes like font size or line spacing tend to have fewer counts than semantic attributes like font or color.
Note that we show the discretized label count for geometric attributes and color.

\begin{table*}[t]
  \centering
  \footnotesize
\caption{Attribute metrics. Acc: accuracy, MAE: mean absolute error, and Diff: CIEDE2000 color difference.}
\vspace{-3mm}
\label{tab:element_fidelity_ablation}
  \begin{tabular}{lccccccccccc}
    \hline
    \multirow{2}{*}{TF}
    &\multirow{2}{*}{Skip}
    & Font 
    & Color
    & Alignment 
    & Capitalization 
    & Font size 
    & Angle 
    & Line spacing
    & Line height 
    \\
    &
    & Acc (\%) $\uparrow$
    & Diff (-) $\downarrow$
    & Acc (\%) $\uparrow$
    & Acc (\%) $\uparrow$
    & MAE (pt) $\downarrow$
    & MAE (°) $\downarrow$
    & MAE (pt) $\downarrow$
    & MAE (-) $\downarrow$
    \\
    \hline
 &\checkmark & 35.3\footnotesize{±0.76} & 53.3\footnotesize{±1.69} & 92.8\footnotesize{±0.68} & 73.5\footnotesize{±1.02} & 23.0\footnotesize{±3.34} & 0.30\footnotesize{±0.06} & 2.15\footnotesize{±0.14} & 0.065\footnotesize{±0.001}  
\\
 \checkmark & & 39.5\footnotesize{±0.48} & 54.0\footnotesize{±0.96} & 93.2\footnotesize{±0.64} & 72.5\footnotesize{±0.75} & 27.7\footnotesize{±0.84} & 0.33\footnotesize{±0.05} & 2.26\footnotesize{±0.15} & 0.092\footnotesize{±0.006} \\
\checkmark & \checkmark & 40.9\footnotesize{±0.76} & 53.7\footnotesize{±1.96} & 93.8\footnotesize{±0.74} & 75.3\footnotesize{±0.67} & 20.9\footnotesize{±0.66} & 0.26\footnotesize{±0.02} & 2.16\footnotesize{±0.16} & 0.065\footnotesize{±0.003}  \\
    \hline
  \end{tabular}
\end{table*}

\begin{table*}[tb]
  \centering
  \footnotesize
\caption{Structure scores (\%).}
\vspace{-3mm}
\label{tab:structure_fidelity_ablation}
  \begin{tabular}{lccccccccccc}
    \hline
    TF
    & Skip
    & Font 
    & Color
    & Alignment 
    & Capitalization
    & Font size 
    & Angle 
    & Line spacing
    & Line height
    \\
    \hline
 &\checkmark & 54.3\footnotesize{±0.66} & 60.1\footnotesize{±0.71} & 64.3\footnotesize{±0.81} & 84.2\footnotesize{±0.78} & 68.0\footnotesize{±0.61}& 84.8\footnotesize{±1.16} & 61.3\footnotesize{±1.18} & 79.1\footnotesize{±0.75}
\\
 \checkmark &  & 67.2\footnotesize{±0.68} & 65.2\footnotesize{±0.36} & 66.0\footnotesize{±0.65} & 86.1\footnotesize{±0.66} & 67.5\footnotesize{±0.58}& 84.1\footnotesize{±1.12} & 62.3\footnotesize{±1.29} & 70.6\footnotesize{±0.49}  \\
\checkmark & \checkmark & 68.6\footnotesize{±0.44} & 66.9\footnotesize{±0.65} & 68.1\footnotesize{±0.58} & 86.3\footnotesize{±0.55} & 71.3\footnotesize{±0.55}& 86.0\footnotesize{±0.37} & 63.8\footnotesize{±0.77} & 78.9\footnotesize{±1.06}\\

    \hline
  \end{tabular}
\end{table*}

\section{Architecture details} \label{sec:arch_detail}

We describe the details of our encoder-decoder architecture in the following.

\paragraph{Encoder}

For each input feature, we project the feature $x_i$ into an embedding $\mathbf{z}_i$ using an encoder: $\mathbf{z}_i = E_i(x_i; \theta)$.
We apply the same encoder to all of the element contexts, where $i$ is an index to the input modalities and elements; i.e., $i=(k, t)$ indicates the $k$-th attribute of the $t$-th element.
For the image feature, we apply ImageNet pre-trained ResNet50~\cite{resnet} to obtain a feature representation.
We apply a pre-trained CLIP~\cite{clip} to encode a text input.
For other categorical features, we apply one-hot encoding to obtain a vector representation.
Once we obtain modality-wise features, we apply a linear projection to all of the features, concatenate all of them into a sequence, and obtain fixed-dimensional embeddings $Z \equiv \{ \mathbf{z}_i \}$.
Let us also denote the set of embeddings belonging to the $t$-th element by $\mathbf{z}_t$ and to the canvas by $\mathbf{z}_\mathrm{canvas}$.

We further apply self-attention transformer modules $F$ to the latent sequence: $Z' \equiv \{ \mathbf{z}'_i \} = F_\mathrm{encoder}(Z; \theta)$ so that the attention mechanism captures any interaction between different modalities across text elements.

\paragraph{Decoder}

We adopt an autoregressive decoder to model the joint distribution of typographic attributes:
\begin{align}
    p_\theta(Y | X) = \prod_t^T p_\theta(\mathbf{y}_t | \mathbf{y}_{t-1}, \dots, \mathbf{y}_1 , X),
\end{align}
and we apply element-wise autoregressive sampling to generate attribute $k$ at the $t$-th element:
\begin{align}
    \hat{y}_{k}^t \sim p_\theta(y_k^t | \mathbf{y}_{t-1}, \dots, \mathbf{y}_1, X).  \label{eq:sampling}
\end{align}

We build the decoder architecture in the following approach:
\begin{align}
    p_\theta(y_k^t | \mathbf{y}_{t-1}, \dots, \mathbf{y}_1, X) &\equiv F_k(\mathbf{h}'_t, \mathbf{s}_t; \theta),\\
    \mathbf{h}'_{t} &= F_\mathrm{decoder}(Z', H_t; \theta),\\
    \mathbf{s}_t &= F_\mathrm{skip}(\mathbf{z}_t, \mathbf{z}_\mathrm{canvas}; \theta).
\end{align}
We model the categorical distribution of each attribute $k$ by the softmax function in the decoder head $F_k$.
Our decoder head takes concatenated features with the outputs from the decoder Transformer $F_\mathrm{decoder}$ and the skip connection $F_\mathrm{skip}(\mathbf{z}_t, \mathbf{z}_\mathrm{canvas})$ which is a shallow MLP.
Our decoder Transformer takes the latent sequence $Z'$ from the encoder and the query sequence $H_t \equiv \{ \mathbf{h}_{1}, \dots, \mathbf{h}_{t} \}$ where:
\begin{align}
    \mathbf{h}_t \equiv \mathbf{p}_t + \sum_{k \in \mathcal{K}} E_k(y_k^{t-1}),
\end{align}
which is a sum of the positional encoding $\mathbf{p}_t$ and additive pooling of the attribute embeddings for $\mathbf{y}_{t-1}$ at the $t$-th text element.
We use the raster scan order of elements, i.e., from top-left to bottom-right, to represent the order of the elements.
$\mathcal{K}$ is a set of typographic attributes for each element.
For $t=1$, we prepare a special \texttt{[start]} token for the second term.
We use the raster scan order, i.e., from top-left to bottom-right, to define the order of elements.

\section{Architecture ablation} \label{sec:arch_ablation}

We ablate the architecture of our model in this section.
We verify the effectiveness of two components the transformer blocks ``TF'' and the skip connection ``Skip''.
Tables~\ref{tab:element_fidelity_ablation} and~\ref{tab:structure_fidelity_ablation} summarize the prediction performance.
We observe that the features from the Transformer blocks improve the prediction of the font and alignment attributes.
While they degrade the performance of prediction in other attributes from the shallow features, i.e., the features from skip-connection, the combined features improve the performance.
These results indicate that both deep features from transformer blocks and shallow features improve the prediction of typographic attributes.
In terms of structure score, the prediction performance through Transformer blocks shows better scores compared to the shallow features in non-geometric attributes and line spacing.
Also, combined features consistently improve the performance except for line spacing.

\section{Additinal qualitative results}
\begin{figure*}[tbp]
\centering
  \includegraphics[width=1.\textwidth]{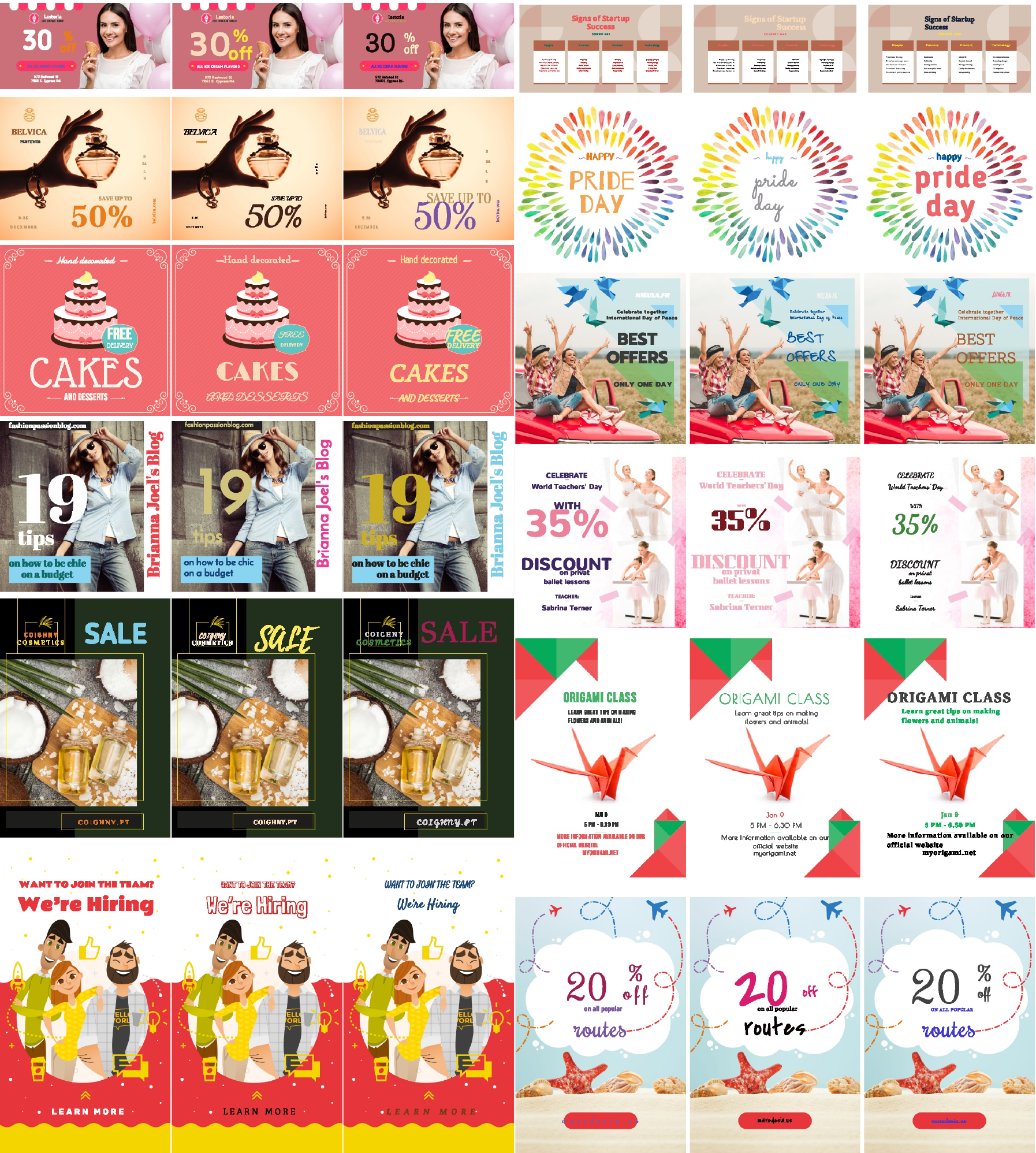}
\caption{Additional diverse generation examples. Each row shows three generated examples for the same input.}
\label{fig:genex_additional}
\end{figure*}

\begin{figure*}[tbp]
\centering
  \includegraphics[width=0.87\textwidth]{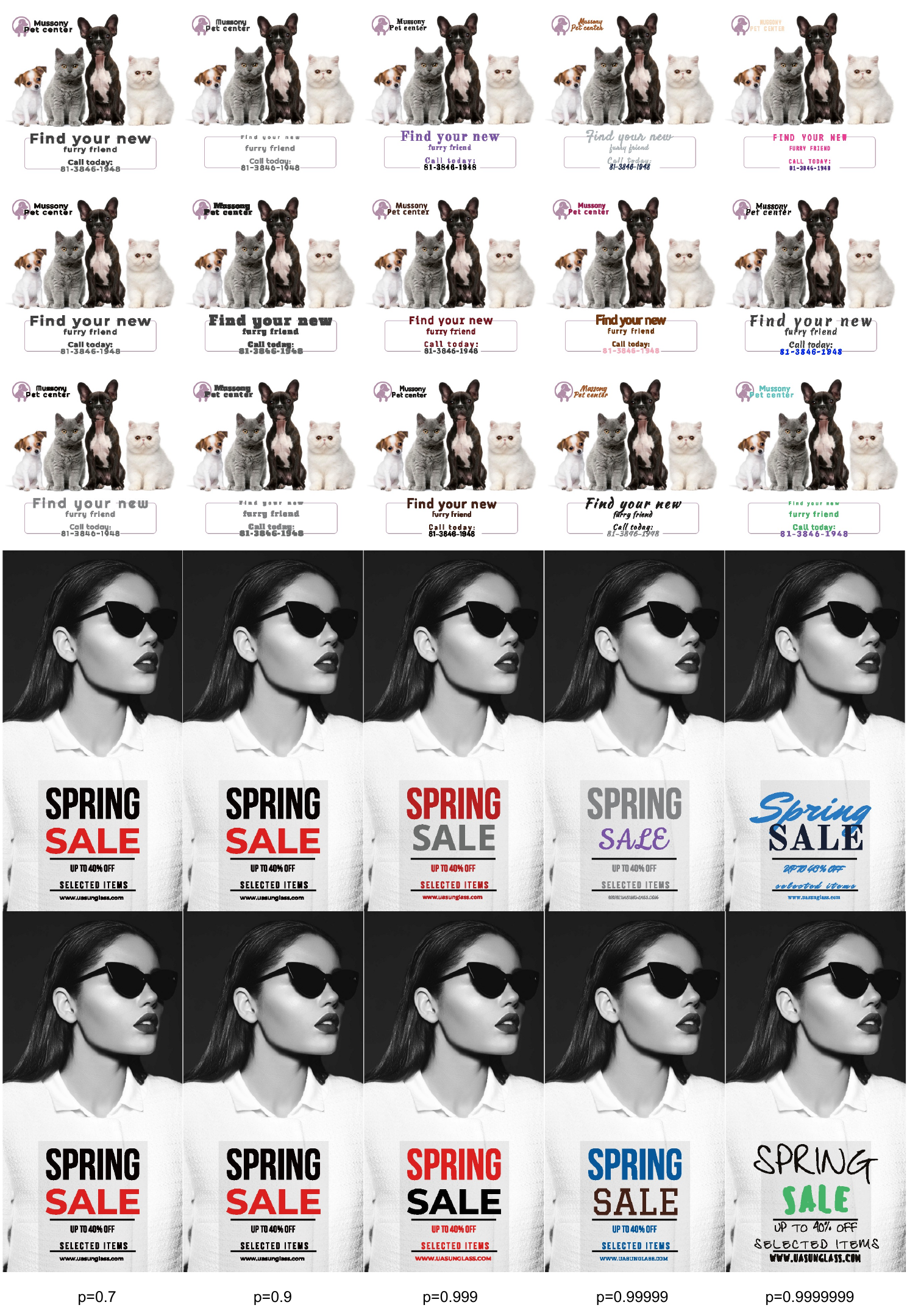}
\caption{Generated examples with different diversity hyper-parameter $p$.}
\label{fig:p_change}
\end{figure*}

Fig.~\ref{fig:genex_additional} shows additional generation examples.
Our model successfully generates appropriate typography in various situations, e.g., many text elements, small text, and large text.
We also show the generated examples with different hyper-parameters $p$ in Fig.~\ref{fig:p_change}.
The sensitivity of hyper-parameters depends on the context.

\end{document}